\documentclass{article}

\usepackage{microtype}
\usepackage{graphicx}
\usepackage{subcaption}
\usepackage{booktabs} 

\usepackage{tcolorbox}
\usepackage{listings}
\usepackage{float}
\usepackage{enumitem}

\usepackage{hyperref}

\usepackage[preprint]{icml2026}

\makeatletter
\renewcommand{\printAffiliationsAndNotice}[1]{\global\icml@noticeprintedtrue}
\makeatother

\usepackage{amsmath}
\usepackage{amssymb}
\usepackage{mathtools}
\usepackage{amsthm}

\usepackage[capitalize,noabbrev]{cleveref}

\theoremstyle{plain}

\theoremstyle{definition}

\theoremstyle{remark}

\usepackage[textsize=tiny]{todonotes}

\icmltitlerunning{CircuitFormer: A Circuit Language Model for Analog Topology Design from Natural Language Prompt}

\begin{document}

\twocolumn[
  \icmltitle{CircuitFormer: A Circuit Language Model for Analog Topology Design from Natural Language Prompt}



  \icmlsetsymbol{equal}{*}

  \begin{icmlauthorlist}
    \icmlauthor{Md Touhidul Islam}{uf}
    \icmlauthor{Sujan Kumar Saha}{uf}
    \icmlauthor{Farimah Farahmandi}{uf}
    \icmlauthor{Mark Tehranipoor}{uf}
  \end{icmlauthorlist}

  \icmlaffiliation{uf}{Department of Electrical and Computer Engineering, University of Florida, Gainesville, FL, USA}

  \icmlcorrespondingauthor{Md Touhidul Islam}{mdtouhidul.islam@ufl.edu}

  \begin{center}
    \textsuperscript{1}University of Florida\\
    \texttt{mdtouhidul.islam@ufl.edu, sujansaha@ufl.edu}\\
    \texttt{farimah@ece.ufl.edu, tehranipoor@ufl.edu}
  \end{center}

  \icmlkeywords{Machine Learning, ICML}

  \vskip 0.3in
]



\printAffiliationsAndNotice{}  

\begin{abstract}
Automating analog circuit design remains a longstanding challenge in Electronic Design Automation (EDA). While Transformer-based Large Language Models (LLMs) have revolutionized software code generation, their application to analog hardware design is hindered by two critical limitations: (i) the scarcity of analog design datasets containing natural language description of a design and its corresponding netlist, and (ii) the inefficiency of general-purpose tokenizers (e.g., Byte Pair Encoding (BPE)) in capturing the inherent graph structure of circuits. To bridge this gap, first, we curate the largest annotated dataset of analog circuit netlists to date, comprising 31,341 netlist-natural language description pairs across all major circuit classes. Furthermore, we propose Circuit Tokenizer (CKT), a novel circuit graph tokenizer designed to encode netlist connectivity by explicitly mining frequent subcircuits. In terms of scalability, CKT overcomes the bottleneck of prior circuit graph serialization methods where vocabulary size scales linearly with maximum number of components in the dataset, $n_{max}$, ($O(n_{\text{max}})$); instead, CKT decouples vocabulary growth from circuit complexity, achieving a constant $O(1)$ complexity. Empirically, CKT outperforms standard BPE on circuit topology representation, reducing sequence length by 57\% and achieving a $2.3\times$ superior compression ratio using a compact, fixed vocabulary of size 512. Leveraging this optimized tokenization, we train a circuit-specific language model, \textbf{CircuitFormer}, a 511M parameter encoder-decoder transformer. Our model achieves 100\% syntactic correctness and an 83\% functional success rate across all major analog circuit categories, outperforming state-of-the-art open-source LLMs by 10\% and 14\%, respectively, while requiring $240\times$ fewer parameters. The dataset is publicly available at \url{https://huggingface.co/datasets/touhid314/cktformer-dataset}.

\end{abstract}

\section{Introduction}
\label{sec:intro}

\begin{table*}[t]
\centering
\caption{Comparison of CircuitFormer with Existing Datasets and Domain-Specific Transformer Models for Analog Design}
\label{tab:combined-comparison}
\begin{small}
\begin{sc}
\resizebox{\textwidth}{!}{%
\begin{tabular}{l c c c c c | c c}
\toprule
& \multicolumn{5}{c|}{\textbf{Dataset Statistics}} & \multicolumn{2}{c}{\textbf{Model Architecture}} \\
\cmidrule(lr){2-6} \cmidrule(l){7-8}
\textbf{Method} & \textbf{Size} & \textbf{\shortstack{Desc.\\Annot.}} & \textbf{\shortstack{Sim.\\Valid.}} & \textbf{\shortstack{Comp.\\Types\textsuperscript{*}}} & \textbf{\shortstack{Auto-\\mated}} & \textbf{Input} & \textbf{Params} \\
\midrule
MasalaChai~\cite{bhandari2025masalachailargescalespicenetlist} & 7,500 & Yes & No & 9 & Yes & \multicolumn{2}{c}{--- \textit{Dataset Only} ---} \\
AMSNet 2.0~\cite{shi2025amsnet20largeams} & 2,686 & No & Yes & \textdagger & No & \multicolumn{2}{c}{--- \textit{Dataset Only} ---} \\
\midrule
AnalogGenie~\cite{gao2025analoggenie} & 3,350 & No & Yes & 8 & No & None & 11.8M \\
AnalogGenie-Lite~\cite{gao2025analoggenielite} & 3,350 & No & Yes & 8 & No & None & 11.8M \\
LaMagic~\cite{pmlr-v235-chang24c} & 2,000 & No & Yes & 6 & Yes & Vector & 248M \\
LaMagic 2~\cite{chang2025lamagic} & 2,000 & No & Yes & 6 & Yes & Vector & 248M \\
\midrule
\textbf{CircuitFormer (Ours)} & \textbf{31,341} & \textbf{Yes} & \textbf{Yes} & \textbf{20} & \textbf{Yes} & \textbf{\shortstack{Natural\\Language}} & \textbf{511M} \\
\bottomrule
\end{tabular}
}
\end{sc}
\end{small}
\begin{flushleft}
    \footnotesize{\textsuperscript{*}Comp. Types: Fundamental SPICE elements (e.g., V, R, L, etc.).}
    \footnotesize{\textdagger Analog Mixed Signal specific}
\end{flushleft}
\end{table*}

The recent success of Large Language Models (LLMs) has transformed software development. Yet analog circuit design remains largely manual and driven by expert heuristics. This analog automation gap persists even though analog circuits are essential, serving as the interface between the physical world and digital computation.

Progress in applying generative AI to analog design is limited by two fundamental challenges. First, high-quality analog circuit data is scarce: unlike software code, most analog designs are buried in textbooks, research papers, or proprietary PDKs. Existing public datasets\cite{bhandari2025masalachailargescalespicenetlist}\cite{shi2025amsnet20largeams} are small, lack natural language descriptions, or cover only narrow classes of circuits. Second, there is a severe \emph{text–graph mismatch}. While standard LLMs operate on linear text, a circuit netlist is simply a textual serialization of an underlying graph. Conventional tokenizers such as BPE are sensitive to component order, even though circuit behavior is invariant to such ordering, leading models to learn superficial text patterns rather than true circuit structure.

Prior work has attempted to address this problem using LLMs\cite{10.1609/aaai.v39i1.32016}\cite{liu2024ampagentllmbasedmultiagentmultistage}, graph neural networks\cite{dong2023cktgnn}, or domain-specific generators\cite{gao2025analoggenie}\cite{pmlr-v235-chang24c}, but each approach has clear limitations: LLMs hallucinate invalid connectivity, graph-based models struggle to align with free-form language, and existing circuit-specific transformers lack scale, control, or natural language conditioning. These limitations motivate a new modeling paradigm.

In this work, we treat circuits as a distinct modality, analogous to images or video in multimodal learning. We introduce \textbf{CircuitFormer}, a 511M-parameter encoder–decoder Transformer designed for translating natural language specifications directly into analog circuit topologies. CircuitFormer is powered by a novel graph-mining Circuit Tokenizer (CKT) that operates on topological subgraphs rather than text strings, allowing the model to focus on structural validity instead of serialization artifacts.

Our contributions are threefold:
\begin{enumerate}[noitemsep, topsep=0pt, leftmargin=*]
\item We release the largest publicly available annotated analog circuit dataset, consisting of 31,341 verified SPICE netlists across all major analog circuit classes.
\item We propose CKT, a domain-specific tokenizer for circuit graphs that exploits graph permutation invariance, reduces sequence length by 57\% compared to general-purpose tokenizers, and achieves $\mathcal{O}(1)$ vocabulary scaling.
\item We train a circuit-specific transformer, CircuitFormer, which achieves 100\% syntactic validity and an 83\% functional success rate on the \textit{CircuitBench-100} benchmark, while being $240\times$ smaller than open-source SOTA LLMs such as GPT-OSS.
\end{enumerate}

We will release the full dataset and source code upon acceptance.

\begin{figure*}[t]
    \centering
    \includegraphics[width=\textwidth]{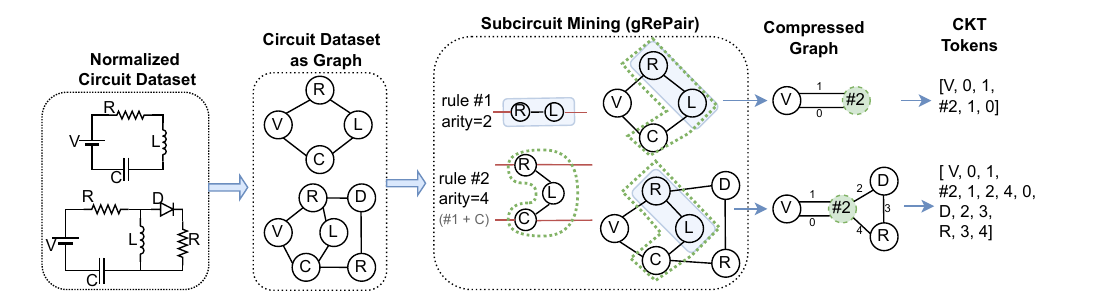}
    \caption{Subcircuit mining and tokenization process.}
    \label{fig:ckt_subckt_mining}
\end{figure*}
\section{Preliminaries and Related Work}
\label{sec:related}

Recent advancements in AI-driven analog circuit design can be broadly categorized into three distinct paradigms: 1) LLM-Based Approaches (Prompt Engineering \& LoRA), 2) Graph Neural Networks, 3) Domain-Specific Transformers.

\textbf{1. LLM-Based Approaches (Prompt Engineering \& LoRA):}
The first paradigm applies off-the-shelf LLMs (e.g., GPT-4, Qwen, DeepSeek) to netlist generation via prompt engineering or parameter-efficient fine-tuning, as in AnalogCoder\cite{10.1609/aaai.v39i1.32016}, AnalogCoder-Pro\cite{lai2025analogcoderprounifyinganalogcircuit}, Artisan\cite{10.1145/3649329.3655903}, AnalogXpert\cite{11100627} and AmpAgent\cite{liu2024ampagentllmbasedmultiagentmultistage}. Despite strong reasoning ability, these models treat netlists as linear text rather than graphs, leading to a fundamental text–graph mismatch. This results in \emph{connectivity hallucinations}, such as invalid or floating nodes that violate circuit laws and fail simulation, and (ii) \emph{inefficient tokenization}, where order-sensitive tokenizers like BPE\cite{sennrich-etal-2016-neural} learn redundant representations for permutation-invariant circuits, wasting model capacity on serialization artifacts instead of structure.

\textbf{2. Graph Neural Networks (GNNs):}
Graph-based approaches such as CktGNN\cite{dong2023cktgnn} operate directly on circuit topology, naturally enforcing permutation invariance and structural validity. However, they struggle with multimodal alignment: mapping free-form natural language prompts to structured graph embeddings remains challenging, limiting their use as flexible, language-driven design agents.

\textbf{3. Domain-Specific Transformers:}
Recent Transformers trained specifically on circuit netlists, including AnalogGenie\cite{gao2025analoggenie} and LaMagic\cite{pmlr-v235-chang24c}, move closer to structure-aware modeling but face key limitations. AnalogGenie is an unconditional generator without user-driven control, while LaMagic relies on rigid, non-natural conditioning formats. Moreover, both are trained on limited-scale datasets, restricting their ability to generalize to complex, real-world analog topologies. AnalogGenie introduces a circuit-specific graph serialization scheme; however, its vocabulary size grows with the maximum number of components in a circuit, which limits scalability and efficiency at larger design scales. 

Table \ref{tab:combined-comparison} provides a comprehensive comparison of recent analog design datasets and domain-specific transformer models.

\subsection{Circuit as a Distinct Modality}
We argue that circuit generation is fundamentally different from code generation: while code is sequential and order-dependent, circuits are physical graphs where component ordering is irrelevant. Accordingly, circuits should be treated as a distinct modality within multimodal learning, analogous to images or video. CircuitFormer operationalizes this view by decoupling topological representation from textual serialization, enabling translation from high-level natural language intent to physically valid, graph-structured circuit designs.

\section{Circuit Tokenizer (CKT)}
\label{ckt_tokenizer}

The CKT tokenizer is a domain-specific tokenization framework designed to represent electronic circuit topologies as compact, invariant sequences suitable for transformer-based architectures. Unlike standard text-based methods such as Byte Pair Encoding (BPE)\cite{sennrich-etal-2016-neural}, and  SentencePiece\cite{kudo-richardson-2018-sentencepiece} which process netlists as raw strings, CKT operates directly on the graphical structure of the circuit. It leverages the Graph Re-Pair (gRePair) algorithm~\cite{MANETH201819} to identify and compress recurring topological substructures.

\subsection{The Tokenization Algorithm}
The CKT algorithm follows a two-stage pipeline: (1) Normalization, (2) Subcircuit Mining.

\subsubsection{Normalization and Instance Anonymization}
The objective of normalization is to distill the netlist into a canonical ``topology-only" representation that retains full electrical connectivity while discarding variable metadata.

\textbf{Canonicalization:} The raw SPICE netlist is first sanitized by removing comments, simulator directives (e.g., \texttt{.model}, \texttt{.tran}), and merging continuation lines. Node identifiers are remapped to a contiguous integer sequence starting from 0, with the global ground (e.g., \texttt{0} or \texttt{gnd}) always mapped to node 0 to preserve topological consistency.

\textbf{Instance Anonymization:} We discard instance names and retain only the device type prefix. For example, instances $R_1$, $R_2$, and $R_{CC}$ are all mapped to a single token type $\langle R \rangle$. This reduces the base vocabulary to the sum of the number of supported SPICE primitives and the maximum number of nodes in a circuit in the dataset. However, a critical challenge of representing a circuit netlist without element names arises with components that reference other devices by name. In SPICE, Current-Controlled Voltage Sources (CCVS, prefix H), Current-Controlled Current Sources (CCCS, prefix F), and Mutual Inductors (prefix K) define their dependencies by referencing the name of a controlling element (typically a voltage source, e.g., \texttt{F1 n1 n2 Vsens}). Since our normalization process strips instance names (reducing \texttt{Vsens} to $\langle V \rangle$), the unique reference to the controlling branch is lost. Furthermore, a branch cannot be uniquely identified solely by its two endpoint nodes if multiple components are connected in parallel between them. However a branch can be uniquely identified with 3 nodes in all cases. Hence, we introduce a Dummy Node Insertion strategy:
\begin{enumerate}[noitemsep, topsep=0pt, leftmargin=*]
    \item We identify any branch that serves as a controlling reference.
    \item We insert a virtual ``dummy" node into that branch, splitting the original edge.
    \item The name-based reference in the dependent source is replaced with a topological reference to this new 3-node sequence (start node, dummy node, end node).
\end{enumerate}

This transformation allows the controlling relationship to be encoded purely through graph connectivity, allowing CKT to represent a circuit without component names and hence reducing the vocabulary space. The overall process after normalization until CKT tokens creation is depicted in Figure \ref{fig:ckt_subckt_mining}.

\subsubsection{Subcircuit Mining via Graph Re-Pair}
CKT employs an iterative graph compression algorithm adapted from Graph Re-Pair (gRePair)~\cite{MANETH201819} to induce a context-free graph grammar by repeatedly collapsing frequent topological substructures into single \emph{macro} tokens.

Training begins by aggregating the entire netlist corpus into a single disjoint union graph, $G_{\text{train}} = \bigsqcup_{c \in \mathcal{D}} G_c$, with unique node prefixes preserving circuit boundaries; this enables global frequency estimation of recurring patterns across the dataset. The algorithm then enumerates all valid \emph{digrams}, defined as pairs of adjacent circuit elements sharing at least one node, while explicitly encoding their connection topology (shared-node mask) to distinguish structures such as series and parallel configurations.

\begin{figure*}[t]
    \centering
    \includegraphics[width=\textwidth]{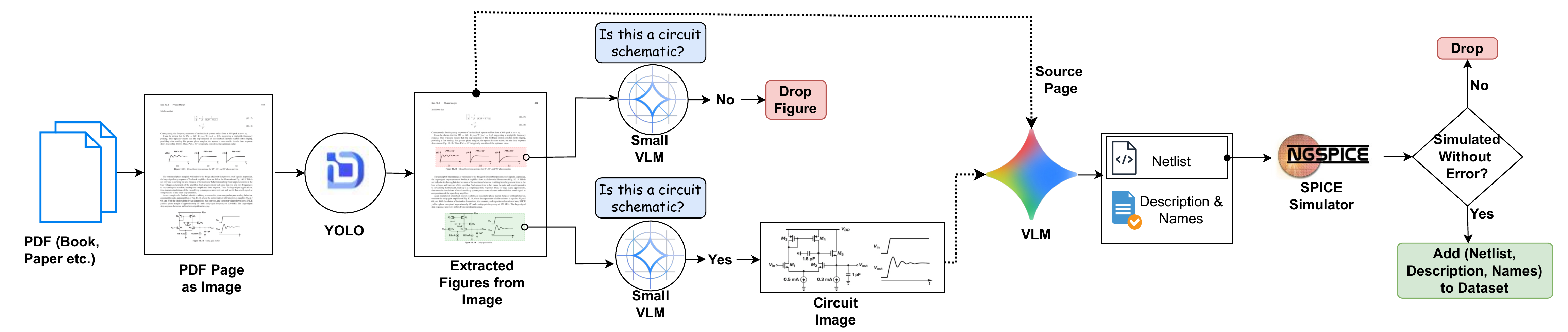}
    \caption{The fully automated dataset making pipeline.}
    \label{fig:dataset_pipeline}
\end{figure*}

   \textbf{Selection via Minimum Description Length (MDL) Gain:}
Unlike frequency-based tokenizers such as BPE or AnalogGenie~\cite{gao2025analoggenie}, CKT selects merge candidates using a gain-based criterion grounded in the Minimum Description Length (MDL) principle. The utility of merging a digram $d$ is defined by its net reduction in representation complexity rather than raw occurrence count. Specifically, the compression gain is computed as
${Gain}(d) = (S_d - 1) \times F_d - S_d$, 
where $S_d$ is the total number of primitive devices spanned by the digram (including primitives within existing macros), and $F_d$ is the number of non-overlapping occurrences in the global graph. The first term captures the savings from removing primitives, while the second accounts for the cost of introducing a new grammar rule. A merge is accepted only if $\text{Gain}(d) > 0$, preventing the vocabulary from being dominated by high-frequency but low-impact patterns.

To ensure efficiency and learnability, we impose two constraints during mining: (i) a minimum frequency threshold $F_{\min}$ to filter rare, non-generalizable patterns, and (ii) a maximum arity $A_{\max}$ that limits the number of external connections of any macro, preventing the creation of overly complex tokens that are difficult to predict or reconstruct.

    \textbf{Rule Creation and Graph Rewriting:}  
After selecting the optimal digram $d^*$, CKT induces a new grammar rule by introducing a non-terminal \emph{macro} token (e.g., \texttt{\#1}, \texttt{\#2}, \dots) that represents the corresponding substructure. The algorithm identifies the external nodes of the digram, those connecting it to the rest of the graph, and defines them as the ports of the macro, preserving the subcircuit’s interface. All subgraphs isomorphic to $d^*$ in $G_{\text{train}}$ are then replaced with this single macro edge, compressing the graph and hierarchically encoding the topology. This process iterates through digram collection, MDL-based selection, and replacement until the target vocabulary size $|\mathcal{V}|$ is reached or no further positive compression gain is possible.

Algorithm \ref{alg:ckt_training} illustrates the complete procedure described above.

\subsection{Complexity Analysis of Vocabulary Size}
\label{ckt-vocab-complexity}
A critical advantage of CKT is that, given a maximum number of nodes allowed, it has $\mathcal{O}(1)$ vocabulary complexity with respect to circuit size in the dataset, resolving a major scalability bottleneck present in prior work. Approaches like AnalogGenie \cite{gao2025analoggenie} and AnalogGenie-Lite \cite{gao2025analoggenielite} rely on instance-specific tokenization, where the vocabulary must explicitly include distinct tokens for every unique identifier (e.g., $R_1, \dots, R_{n_{max}}$).

\begin{algorithm}[H]
   \caption{Circuit Tokenizer (CKT) Training Procedure}
   \label{alg:ckt_training}
\begin{algorithmic}
   \STATE {\bfseries Input:} Netlist Corpus $\mathcal{D}$, Target Vocabulary Size $V_{max}$, Min Frequency $F_{min}$, Max Arity $A_{max}$, Min Gain $G_{min}$
   \STATE {\bfseries Output:} Grammar Rules $\mathcal{R}$, Tokenized Graphs
   \STATE
   \STATE \COMMENT{\textit{Phase 1: Normalization \& Graph Construction}}
   \STATE $G_{train} \leftarrow \emptyset$
   \FORALL{netlist $N \in \mathcal{D}$}
       \STATE $N' \leftarrow$ \textsc{Normalize}($N$) \COMMENT{Remove comments, standardize names}
       \STATE $N'' \leftarrow$ \textsc{AddDummyNodes}($N'$) \COMMENT{Handle branch references (Sec \ref{ckt_tokenizer})}
       \STATE $G_{train} \leftarrow G_{train} \cup \textsc{ToGraph}(N'')$ \COMMENT{Disjoint union}
   \ENDFOR
   \STATE
   \STATE \COMMENT{\textit{Phase 2: Subcircuit Mining (gRePair)}}
   \STATE $\mathcal{R} \leftarrow$ Initial Primitive Tokens
   \WHILE{$|\mathcal{R}| < V_{max}$}
       \STATE $\mathcal{M} \leftarrow$ \textsc{EnumerateDigrams}($G_{train}$)
       \STATE $d^* \leftarrow \text{None}, \quad \text{MaxGain} \leftarrow 0$
       \FORALL{digram $d \in \mathcal{M}$}
           \STATE $S_d \leftarrow$ \textsc{GetSpan}($d$)
           \STATE $F_d \leftarrow$ \textsc{CountFrequency}($d, G_{train}$)
           \STATE $A_d \leftarrow$ \textsc{GetArity}($d$) \COMMENT{Number of external ports}
           
           \IF{$F_d < F_{min}$ \OR $A_d > A_{max}$}
               \STATE \textbf{continue} \COMMENT{Skip invalid candidates}
           \ENDIF
           
           \STATE $Gain \leftarrow (S_d - 1) \times F_d - S_d$ \COMMENT{MDL Objective}
           \IF{$Gain > \text{MaxGain}$}
               \STATE $d^* \leftarrow d, \quad \text{MaxGain} \leftarrow Gain$
           \ENDIF
       \ENDFOR
       
       \IF{$\text{MaxGain} < G_{min}$}
           \STATE \textbf{break} \COMMENT{Gain threshold not met}
       \ENDIF
       
       \STATE $r_{new} \leftarrow \textsc{CreateRule}(d^*)$
       \STATE $G_{train} \leftarrow \textsc{ReplaceOccurrences}(G_{train}, d^*, r_{new})$
       \STATE $\mathcal{R} \leftarrow \mathcal{R} \cup \{r_{new}\}$
   \ENDWHILE
   \STATE \textbf{return} $\mathcal{R}$
\end{algorithmic}
\end{algorithm}

Consequently, their vocabulary size $|\mathcal{V}|$ scales linearly as $\mathcal{O}(n_{max})$, where $n_{max}$ is the maximum component count in the dataset, leading to vocabulary explosion and an inability to generalize to circuits larger than those seen during training. In contrast, by decoupling topology from instance naming (mapping all $\{R_1, \dots, R_k\} \to \langle R \rangle$) and defining a fixed limit for learned macro-rules, CKT maintains a constant vocabulary size for a maximum number of nodes allowed.

\subsection{Detokenization}
\label{detokenization_sec}

The detokenization process reconstructs the full circuit topology by recursively expanding CKT macro tokens into their underlying primitive components and resolving node connectivity. Because grammar rules explicitly encode shared-node masks and external ports, this inverse transformation is lossless. A detailed description of the detokenization algorithm is provided in Appendix \ref{detokenization}.

\subsection{Line Shuffle Augmentation}
\label{sec:line-shuffle}
Because CKT identifies components based on their graph-domain connections rather than their position in the text, we can randomly shuffle the lines without affecting how well the tokenizer compresses the circuit. This allows us to effectively do line shuffle augmentation during training - we randomly shuffle the lines every time the model sees an example. A circuit with $n$ lines has $n!$ valid permutations, meaning there are millions of ways to write the exact same circuit. This prevents the model from memorizing a specific text order and forces it to learn the actual circuit topology.

\section{The CircuitFormer Dataset}
\label{sec:dataset}


To enable the training of large-scale generative models for analog design, we curated the largest dataset of analog circuit topologies to date. Our dataset comprises 31,341 unique SPICE netlists. Each netlist is paired with a natural language description, a list of possible names, and the original schematic image. The data was sourced from 62 prominent textbooks and reference volumes in analog integrated circuit design. Every netlist in the dataset has been verified via Ngspice simulation to ensure syntactical validity.

Unlike most previous methods, our approach is fully automated. It accurately extracts netlists from diverse PDF sources, including textbooks and research papers. This scalable design allows to easily expand the dataset. Furthermore, the system automatically assigns circuits with their correct names and descriptions, removing the need for manual annotation.



\subsection{Automated Dataset Creation Pipeline}
\label{sec:pipeline}
Due to the scarcity of open-source, high-quality SPICE netlists online, we identified textbooks as the optimal source for high-fidelity circuit data. However, converting static PDF pages into simulatable code requires a robust extraction pipeline. Our automated workflow consists of three stages: 

\textbf{1. Visual Extraction and Filtering:}
We first rasterize the textbook PDFs into high-resolution images. To isolate circuit diagrams from text and other figures, we employ the DocLayout-YOLO\cite{zhao2024doclayoutyoloenhancingdocumentlayout} object detection model. Extracted figures are then passed through a lightweight Vision-Language Model (VLM) filter to discard non-circuit images (e.g., block diagrams, signal plots).

\textbf{2. Context-Aware Netlist Parsing:}
We utilize Gemini 3 to parse the identified schematic images into SPICE netlists. Along with the cropped schematic image, we provide the VLM with the full source page containing the figure. This contextual grounding allows the model to accurately extract the circuit's functional description and canonical name from the caption and surrounding text, and to resolve ambiguous component values or model parameters that are defined in the text but omitted from the schematic.

\textbf{3. Simulation-Based Verification:}
The raw netlists generated by the VLM are sanitized and executed in the Ngspice simulator. Any netlist that fails to converge or produces syntax errors is automatically discarded. This rigorous filtering ensures that our dataset contains only electrically valid, syntactically correct circuits.

Figure \ref{fig:dataset_pipeline} shows the full dataset making process.

\subsection{Can Vision Language Models accurately Parse Netlists from Circuit Schematic Images?}
Our adoption of VLMs for netlist extraction is grounded in the observation that very recent state-of-the-art models exhibit remarkable proficiency in translating circuit schematic images into valid SPICE netlists. To validate the feasibility of this approach, we assessed the capabilities of these models in converting circuit schematic images into structured netlists. Appendix \ref{app:vlm-ckt2nl} provides detailed experimental evidence demonstrating that Gemini 3 performs this task with notably high accuracy and reliability.

To rigorously verify the quality of the dataset generated via this approach, we performed a manual validity audit on the final dataset. We randomly sampled 50 instances from the final dataset and manually inspected the VLM-generated netlists against the source images. As shown in Table \ref{tab:dataset-validity-test}, \textbf{CircuitFormer} (utilizing the Gemini 3-generated ground truth) achieved a 95\% perfect image-to-netlist performance. The single failure case involved a minor connectivity error affecting only one element, resulting in an effective element-wise accuracy of 99.25\%.


\subsection{Dataset Statistics}
The dataset covers a comprehensive range of 20 major SPICE element types, excluding only the rare A, O, P, U, W, and Y elements. In total, it comprises 328,565 SPICE components. The complexity of the circuits varies widely, ranging from single-component primitives to large systems with up to 139 components. Detailed statistical distributions of the dataset are provided in Appendix \ref{app:dataset-stats}.
\begin{table}[h]
\centering
\caption{Manual validity audit of random samples from the dataset ($N=50$).}
\label{tab:dataset-validity-test}
\resizebox{\columnwidth}{!}{%
\begin{small}
\begin{sc}
\begin{tabular}{l c c}
\toprule
Method & Netlist Accuracy & Element Accuracy \\
\midrule
Masala Chai & 6.0\% & 36.92\% \\
\textbf{CircuitFormer (Ours)} & \textbf{95.0\%} & \textbf{99.25\%} \\
\bottomrule
\end{tabular}
\end{sc}
\end{small}
}
\vspace{-5mm}
\end{table}

\section{Results}
\label{sec:results}

\subsection{Experimental Setup}

\textbf{The ``CircuitBench-100" Test Set:}
Due to the absence of standardized benchmarks for text-to-netlist generation, we curated \textit{CircuitBench-100}, a novel test set consisting of 100 natural language prompts. The set spans diverse topologies including single and multi-stage amplifiers, active/passive filters, oscillators, and power management ICs (PMICs). Each entry includes a high-level design prompt and a ground-truth netlist for reference. Detailed statistics on the topological diversity of this benchmark are provided in Appendix \ref{sec:appendix-testset}.

\textbf{Dataset and Preprocessing:}
From the curated corpus of 31,341 netlists, we filtered the training set to exclude netlists with subcircuit (`X') calls. This ensures that the model learns fundamental device physics rather than hierarchical abstractions. The resulting dataset of 18,011 valid netlists was split into a 94\% training set and a 6\% validation set ($\approx 1,000$ samples). To prevent overfitting and enforce topological invariance, we applied the line-shuffling augmentation technique described in Section \ref{sec:line-shuffle} during training. 

\textbf{Training Setup:}
\begin{figure}[t]
    \centering
    \includegraphics[width=0.6\columnwidth]{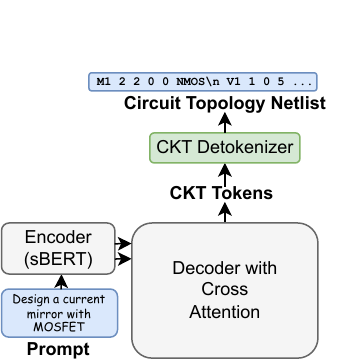}
    \caption{The Overall Transformer architecture used.}
    \label{fig:model-architecture}
\vspace{-6mm}
\end{figure}
\textbf{CircuitFormer} is a hybrid encoder–decoder Transformer that uses a 110M parameter MPNet encoder \cite{song2020mpnetmaskedpermutedpretraining} to capture the semantic meaning of natural language prompts and a GPT-2-based decoder\cite{radford2019language} extended with cross-attention layers. We use 24-layers with $d_{model}=1020$ and $n_{heads}=12$ for the decoder. The encoder uses a wordpiece tokenizer while the decoder uses the CKT tokenizer. The overall model architecture is presented in Figure \ref{fig:model-architecture}. Training is done in three stages: (1) pre-train the decoder on netlists to learn circuit syntax(100 epochs, $5 \times 10^{-5}$ learning rate), (2) align the encoder and decoder by training only the cross-attention layers for 25 epochs, and (3) fine-tune the full model end-to-end to jointly optimize semantic understanding and topology generation (75 epochs, $2 \times 10^{-5}$ learning rate). AdamW optimizer with a 10\% linear warmup was used.
For inference, we employed beam search with a beam width of $k=4$ to ensure generation consistency. 
All experiments were conducted on a single NVIDIA H100 GPU with 96GB VRAM. 

We apply line-shuffle augmentation (Section \ref{sec:line-shuffle}) during training to encourage the model to learn the underlying circuit graph structure rather than relying on a fixed textual ordering.

Appendix \ref{sec:model-architecture} contains details about the model architecture and the training strategy.

\textbf{Baselines:}
We compare CircuitFormer against state-of-the-art Open-Source Large Language Models (LLMs). Domain-specific transformers like AnalogGenie were excluded from direct comparison as they lack natural language conditioning capabilities. Consequently, open-source LLMs represent the only viable baseline for diverse, prompt-based analog generation. We evaluate: \textit{GPT-OSS (120B)}\cite{openai2025gptoss120bgptoss20bmodel}, \textit{Llama 3.3 (70B)}\cite{grattafiori2024llama3herdmodels}, \textit{Mistral Large 2 (123B)}\cite{mistral2024large2}, \textit{Gemma 3 (27B)}\cite{gemmateam2025gemma3technicalreport}, and \textit{DeepSeek-V3.2 (685B)}\cite{deepseekai2025deepseekv32pushingfrontieropen}.

\textbf{Circuit Generation Evaluation Metrics:}
\begin{enumerate}[noitemsep, topsep=0pt, leftmargin=*]
    \item \textbf{Success Rate:} Since valid analog circuit designs are non-unique, traditional metrics like exact string matching or graph isomorphism are insufficient for evaluating functional correctness. Consequently, we rely on manual human expert evaluation to determine whether a generated topology satisfies the design prompt, using the ground-truth netlist as a functional reference. While manual expert inspection provides the highest fidelity, it is prohibitively expensive and difficult to scale or reproduce. To address this, we also implemented an automated evaluation pipeline using \textit{Gemini 3} as an LLM Judge. Appendix \ref{app:judge-llm} contains detail about the judge LLM.
       
    \item \textbf{Validity Rate:} The percentage of generated netlists that pass Ngspice simulation without syntax errors or convergence failures.
    
    \item \textbf{Maximum Mean Discrepancy (MMD):} We calculate the MMD between the graph of the generated topologies and the ground truth topologies in the test set. 
\end{enumerate}

\textbf{Tokenizer Evaluation Metrics:}
\begin{enumerate} 
    \item \textbf{Compression Ratio (CR):} Defined as the ratio of characters in the raw topology to the number of tokens produced:
    \begin{equation}
        \text{CR} = \frac{\text{Total Characters (Topology-Only)}}{\text{Total Tokens}}
    \end{equation}
    Note: To ensure a fair comparison, ``Topology-Only" inputs strip variable values and model names for both CKT and baseline tokenizers.
    \item \textbf{Sequence Length Reduction (SLR):} The percentage reduction in sequence length achieved by CKT compared to other baseline tokenizer:
    \begin{equation}
        \text{SLR} = \left( 1 - \frac{\text{Avg Tokens}_{CKT}}{\text{Avg Tokens}_{Baseline}} \right) \times 100
    \end{equation}
\end{enumerate}

\subsection{Evaluation of CKT Tokenizer}

\begin{figure}[ht]
  \vskip 0.2in
  \begin{center}
    \centerline{\includegraphics[width=\columnwidth]{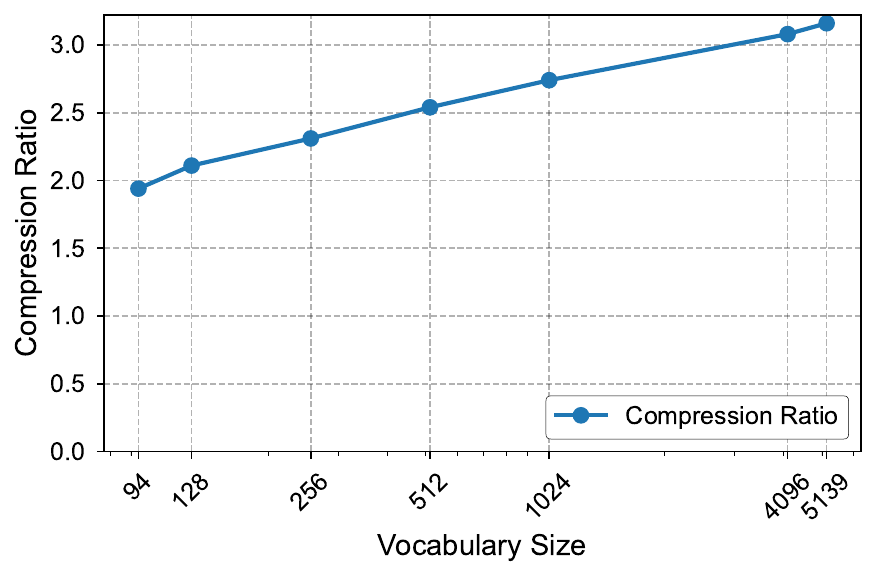}}
    \caption{Impact of vocabulary size on compression ratio. The ``Min Vocab" (94) represents the base primitive set, while 5139 represents the maximal rule set mined with $F_{min}=2$. Higher vocabulary sizes yield denser latent representations.}
    \label{fig:compress-stats}
  \end{center}
  \vskip -0.2in
  \vspace{-4mm}
\end{figure}

\textbf{Compression Efficiency:}
We first analyze the impact of vocabulary size on compression efficiency. As shown in Figure \ref{fig:compress-stats}, increasing the vocabulary size via subgraph mining directly correlates with higher compression. Table \ref{tab:tokenizer-comparison} benchmarks CKT against standard text tokenizers. Even with the minimal vocabulary of 92 tokens (representing the base primitive set with no grammar rules added), CKT achieves a 43.8\% reduction in sequence length compared to GPT-4's `tiktoken'. When applying subcircuit mining under the constraints of $F_{min}=2$, $Gain_{min}=1$, and $A_{max}=4$, the tokenizer reaches a maximum vocabulary size of 5139. At this maximum compression level, CKT achieves a massive 65.5\% reduction, demonstrating that topological macro-tokens are significantly more information-dense than character-based BPE tokens.

\begin{table*}[t]
\centering
\caption{Comparison of Tokenizer Compression and Sequence Length Reduction (Topology-Only Inputs)}
\label{tab:tokenizer-comparison}
\begin{small}
\begin{sc}
\begin{tabular}{llcccc}
\toprule
Tokenizer & Implementation & \begin{tabular}[c]{@{}c@{}}Compression\\ Ratio\end{tabular} & Vocab Size & \begin{tabular}[c]{@{}c@{}}Reduction vs\\ BPE (\%)\end{tabular} & \begin{tabular}[c]{@{}c@{}}Reduction vs\\ SentencePiece (\%)\end{tabular} \\ \midrule
BPE & Tiktoken (GPT-4) & 1.09 & 200k & --- & --- \\
SentencePiece & Gemma 3 & 1.04 & 262k & --- & --- \\
CKT (Ours) & Min Vocab & 1.94 & 92 & 43.8\% & 46.4\% \\
CKT (Ours) & Optimal Vocab & 2.54 & 512 & 57.1\% & 59.1\% \\
CKT (Ours) & Max Vocab & \textbf{3.16} & 5139 & \textbf{65.5\%} & \textbf{67.1\%} \\ \bottomrule
\end{tabular}
\end{sc}
\end{small}
\end{table*}

\subsection{Evaluation of Circuit Generation}

\textbf{Comparison with State-of-the-Art:}
Table \ref{tab:model-comparison} presents the comparative results on \textit{CircuitBench-100}. CircuitFormer outperforms massive general-purpose models by a wide margin. While DeepSeek-V3.2 (685B) achieves a competitive 94\% validity rate, it only reaches a 63\% success rate in functional design. In contrast, CircuitFormer achieves 100\% validity and an 83\% success rate, demonstrating that domain-specific tokenization and training allows a model $1000\times$ smaller to master circuit topology.

\begin{table*}[t]
  \caption{Comparison of CircuitFormer with SOTA Open Source LLMs on CircuitBench-100.}
  \label{tab:model-comparison}
  \begin{center}
    \begin{small}
      \begin{sc}
        \begin{tabular}{lcccc}
          \toprule
          Model & Params & Success Rate ($\uparrow$) & Valid Rate ($\uparrow$) & MMD \\
          \midrule
          GPT-OSS & 120B & 69\% & 90\% & 0.180 \\
          Llama 3.3 & 70B & 31\% & 60\% & 0.224 \\
          Mistral Large 2 & 123B & 32\% & 85\% & 0.219 \\
          Gemma 3 & 27B & 23\% & 64\% & 0.186 \\
          DeepSeek V3.2 & 685B & 63\% & 94\% & 0.202 \\
          \midrule
          \textbf{CircuitFormer (Ours)} & \textbf{0.5B} & \textbf{83\%} & \textbf{100\%} & 0.232 \\
          \bottomrule
        \end{tabular}
      \end{sc}
    \end{small}
  \end{center}
  \vskip -0.1in
\end{table*}

\begin{figure}[ht]
  \vskip 0.2in
  \begin{center}
    \centerline{\includegraphics[width=\columnwidth]{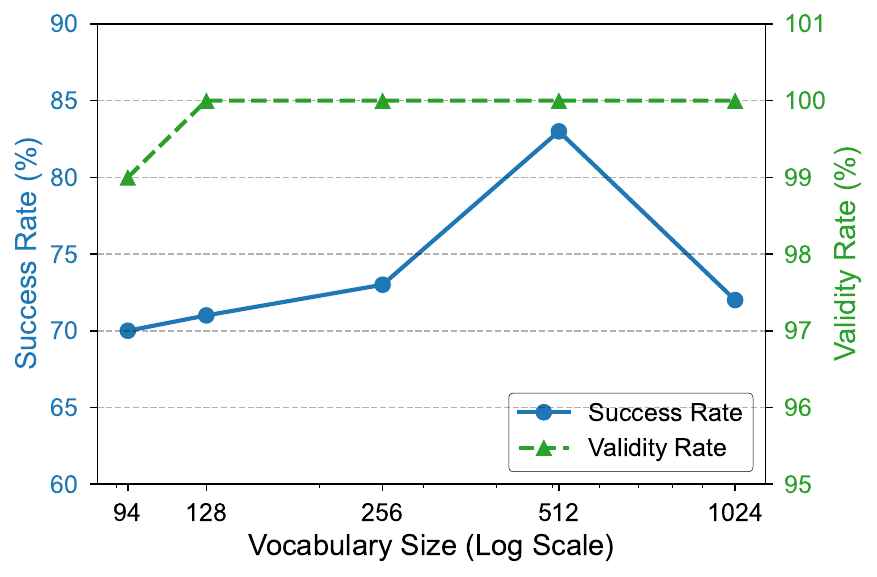}}
    \caption{Impact of vocabulary size on generation quality. While validity remains robust at 100\%, success rate peaks at $|\mathcal{V}|=512$, highlighting the trade-off between sequence compression and token sparsity.}
    \label{fig:vocab-perform}
  \end{center}
  \vskip -0.2in
  \vspace{-5mm}
\end{figure}

\textbf{Impact of Vocabulary Size on Generation:}
We analyzed how vocabulary size affects model performance (Figure \ref{fig:vocab-perform}) and observed an inverted-U trend. With a small vocabulary (under 100 tokens), the success rate is limited to 70\% because the resulting sequences are too long for the model to handle effectively. Performance improves as we add more tokens, reaching a peak of 83\% success at around 512 tokens. This is the ``sweet spot" where we balance shorter sequences with common, easy-to-learn tokens. However, beyond 1024 tokens, performance starts to drop. This happens because many subcircuit tokens become too rare in the training data, making it difficult for the model to learn them well.

\textbf{Generalization vs. Memorization:}
CircuitFormer exhibits a slightly higher MMD (0.232) compared to baselines like GPT-OSS. In the context of our 100\% validity rate, this is a positive indicator. A low MMD often suggests the model is reproducing ``average" circuits close to the training distribution. The higher MMD, coupled with high functional success, suggests that CircuitFormer has generalized the underlying design rules and is synthesizing novel topological variations that differ distributionally from the training set while maintaining electrical correctness.

\subsection{Ablation Studies}
We conducted an ablation study on the CKT tokenizer and the line shuffle augmentation strategy to understand their overall impact. Table \ref{tab:ablation} summarizes the results of our ablation studies.

\textbf{Impact of Graph-Aware Tokenization (CKT vs. BPE):}
To quantify the benefit of the CKT tokenizer over a general-purpose tokenizer in the overall system, we trained a variant of CircuitFormer using a standard BPE tokenizer instead of CKT. The BPE tokenizer was trained on the same training dataset with a vocabulary size of 2000, while keeping all other settings identical. As shown in Table \ref{tab:ablation}, replacing CKT with BPE results in a sharp decline in performance. The BPE-based model achieves only a 71\% success rate, compared to 83\% for the CKT-based model, even though CKT uses a 74.4\% smaller vocabulary.

\textbf{Impact of Line Shuffle Augmentation:}
We also evaluated the importance of our line-shuffling augmentation strategy. We can see in table \ref{tab:ablation} that just turning off the line shuffle augmentation, drops the success rate to 23\% while maintaining a validity score of 98\%. These results highlight the critical role of line-shuffle augmentation in improving training effectiveness and overall success.

\begin{table}[h]
\centering
\caption{Ablation Study Results. Both the CKT tokenizer and Line Shuffle augmentation are critical for achieving high functional success.}
\label{tab:ablation}
\begin{small}
\begin{sc}
\begin{tabular}{lcc}
\toprule
\textbf{\shortstack{Model\\Variant}} & \textbf{Validity (\%)} & \textbf{\shortstack{Success\\Rate (\%)}} \\
\midrule
\textbf{CircuitFormer (Full)} & \textbf{100\%} & \textbf{83\%} \\
w/o CKT (Standard BPE) & 98\% & 70\% \\
w/o Line Shuffle & 98\% & 23\% \\
\bottomrule
\end{tabular}
\end{sc}
\end{small}
\vspace{-5mm}
\end{table}

\section{Conclusion and Future Scope}
\label{sec:conclusion}

We introduced CircuitFormer and the Circuit Tokenizer (CKT), establishing the first framework capable of generating valid analog topologies directly from natural language by resolving the fundamental text-graph mismatch. Beyond analog design, CKT’s ability to decouple topology from serialization offers a scalable, permutation-invariant solution for other graph-structured domains, such as molecular discovery and network analysis. Furthermore, this methodology holds significant promise for automating direct transistor-level digital synthesis, paving the way for a unified, end-to-end generative design flow across the entire hardware spectrum.

\section{Impact Statement}

This work addresses the critical data scarcity bottleneck in Electronic Design Automation (EDA) by releasing the largest annotated analog circuit dataset to date, a contribution likely to catalyze a new wave of data-driven research in analog automation. Beyond hardware, the proposed Circuit Tokenizer (CKT) establishes a scalable, permutation-invariant framework for graph serialization, offering a transferable methodology with significant potential for advancing generative modeling in other graph-intensive domains such as chemoinformatics, social network analysis, and knowledge representation.

\bibliography{paper}
\bibliographystyle{icml2026}

\newpage
\appendix
\onecolumn

\clearpage
\section{Example of Circuits Designed by CircuitFormer}

\begin{figure*}[h]
    \centering
    \begin{subfigure}[b]{0.45\textwidth}
        \centering
        \fbox{\parbox{0.9\textwidth}{
            \textbf{Prompt:} \\
            "Create a 3-stage ring oscillator using CMOS inverters".
        }}
        \caption{Input Prompt}
        \label{fig:oscillator_prompt}
    \end{subfigure}
    \hfill 
    \begin{subfigure}[b]{0.45\textwidth}
        \centering
        \includegraphics[width=\linewidth]{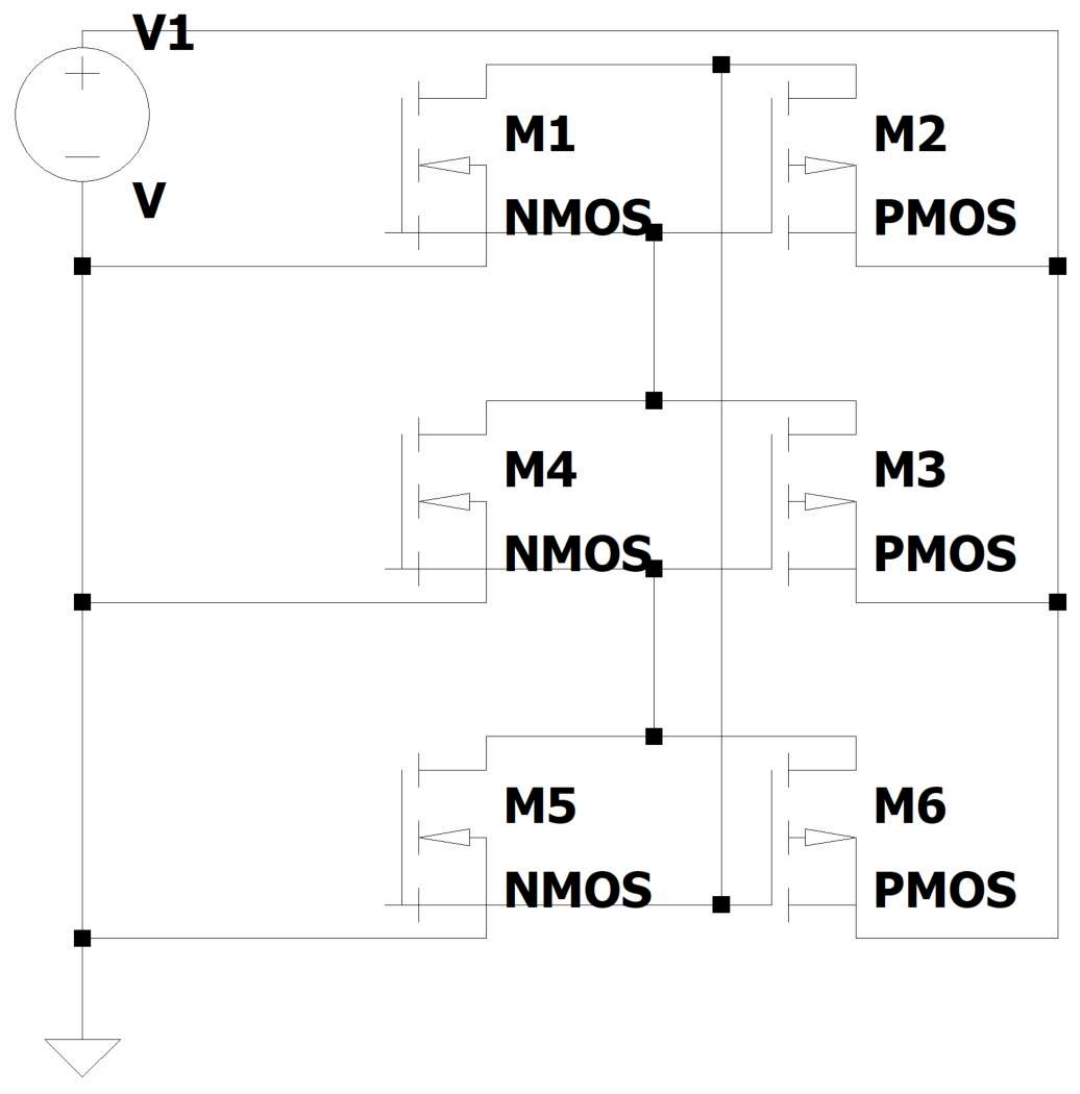}
        \caption{Generated Circuit Topology}
        \label{fig:oscillator_circuit}
    \end{subfigure}
    
    \caption{CircuitFormer generating a ring oscillator from a natural language prompt. (a) The input user prompt. (b) The corresponding valid CMOS topology generated by CircuitFormer.}
    \label{fig:qualitative_result}
\end{figure*}

\begin{figure*}[h]
    \centering
    \begin{subfigure}[b]{0.45\textwidth}
        \centering
        \fbox{\parbox{0.9\textwidth}{
            \textbf{Prompt:} \\
            "design a bjt differential amplifier with an active load current mirror and a tail current source".
        }}
        \caption{Input Prompt}
        \label{fig:oscillator_prompt}
    \end{subfigure}
    \hfill 
    \begin{subfigure}[b]{0.45\textwidth}
        \centering
        \includegraphics[width=\linewidth]{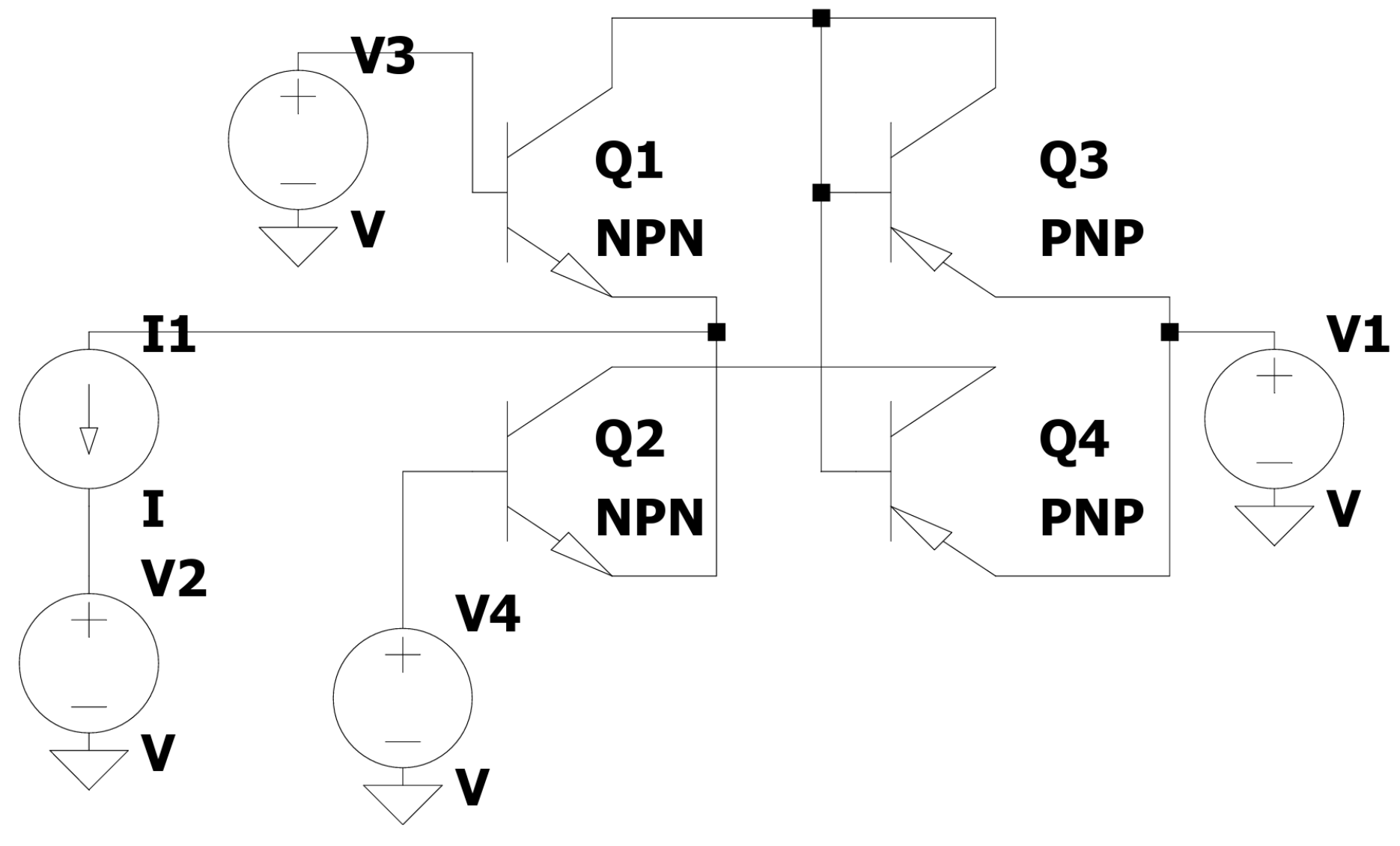}
        \caption{Generated Circuit Topology}
        \label{fig:oscillator_circuit}
    \end{subfigure}
    
    \caption{CircuitFormer generating a differential amplifier from a natural language prompt. (a) The input user prompt. (b) The corresponding valid topology generated by CircuitFormer.}
    \label{fig:qualitative_result}
\end{figure*}

\clearpage
\section{CKT Detokenization and Reconstruction Algorithm}
\label{detokenization}

The detokenization process serves as the inverse operation to CKT tokenization, strictly designed to be a lossless transformation with respect to circuit topology. The algorithm reconstructs a valid, simulatable SPICE netlist from a flat sequence of CKT tokens through a two-stage pipeline: (1) Recursive Graph Expansion, which recovers the primitive graph topology, and (2) Topological Reconstruction, which resolves semantic dependencies and SPICE syntax requirements.

\subsection{Stage 1: Recursive Macro Expansion (Graph-Level)}
The objective of this stage is to "unroll" the hierarchical grammar rules learned during training to recover the fundamental graph of primitive components.

\textbf{Queue-Based Expansion:}
The algorithm initializes a First-In-First-Out (FIFO) queue with the input token sequence. It operates iteratively:
\begin{enumerate}
    \item \textbf{Dequeue:} An edge (token) $e$ is popped from the queue.
    \item \textbf{Check Type:}
    \begin{itemize}
        \item If $e$ is a \textbf{Primitive} (e.g., $\langle R \rangle$, $\langle M \rangle$): It is added to the final set of resolved components.
        \item If $e$ is a \textbf{Macro} (e.g., $\langle \#25 \rangle$): The algorithm retrieves the corresponding Grammar Rule $r$ that generated this macro.
    \end{itemize}
    \item \textbf{Pattern Mapping:} The rule $r$ defines the connectivity between the macro's external ports ($p_0, p_1, \dots$) and its two constituent children (Left $L$ and Right $R$). The algorithm maps the concrete node IDs of the current edge $e$ to the rule's internal placeholders to generate two new concrete edges, $e_L$ and $e_R$.
    \item \textbf{Recurse:} The new edges $e_L$ and $e_R$ are pushed back into the queue for further processing. This allows for deep hierarchical structures to be unrolled level-by-level until only primitives remain.
\end{enumerate}

\subsection{Stage 2: Topological Reconstruction (SPICE-Level)}
Once the primitive graph is recovered, the node identifiers are normalized integers. The second stage transforms this raw graph into valid SPICE syntax, reversing the "Topological Branch Identification" transformations applied during normalization.

\textbf{1. Branch Re-merging (Controlling Sources):}
During normalization, branch-defining elements (Voltage Sources $V$, Inductors $L$) were split into 3-node sequences using a "dummy" node to allow for topological referencing. The reconstruction algorithm scans for these specific topological signatures (e.g., a pair of edges identifying as partial split-branches). It merges the 3-node branch back into a single 2-node component, effectively removing the dummy node and restoring the original branch.

\textbf{2. Reference Resolution ($F, H, K$ Elements):}
Dependent sources (CCCS $F$, CCVS $H$) and mutual inductors ($K$) rely on named references to controlling branches. In the raw graph, these dependencies exist as direct connections to the (now-resolved) dummy nodes.
The algorithm traverses the graph to identify components connected to these resolved branches. It replaces the explicit node connection with the \textit{name} of the restored controlling component (e.g., converting a topological connection at node $n_{dummy}$ into the text string \texttt{"Vsense"}).

\textbf{3. Canonical Naming and Value Injection:}
Since instance names ($R_1, R_2$) are abstracted during tokenization, the reconstructor assigns stable, canonical instance names based on component type and traversal order (e.g., $R\_0, R\_1, M\_0 \dots$).

The complete detokenization procedure, integrating both the recursive graph expansion and the topological reconstruction, is summarized in Algorithm \ref{alg:detokenization}.

\begin{algorithm}
   \caption{CKT Detokenization Procedure}
   \label{alg:detokenization}
\begin{algorithmic}
   \STATE {\bfseries Input:} Token Sequence $T$, Grammar $\mathcal{R}$
   \STATE {\bfseries Output:} SPICE Netlist $N_{out}$
   \STATE
   \STATE \COMMENT{\textit{Stage 1: Recursive Expansion}}
   \STATE $Q \leftarrow$ InitializeQueue($T$)
   \STATE $E_{primitives} \leftarrow \emptyset$
   \WHILE{$Q$ is not empty}
       \STATE $e \leftarrow Q.\text{pop}()$
       \IF{\text{IsPrimitive}($e$)}
           \STATE $E_{primitives}.\text{add}(e)$
       \ELSE
           \STATE $Rule \leftarrow \mathcal{R}[\text{GetID}(e)]$
           \STATE $e_L, e_R \leftarrow \text{MapNodes}(e, Rule)$
           \STATE $Q.\text{push}(e_L, e_R)$
       \ENDIF
   \ENDWHILE
   \STATE
   \STATE \COMMENT{\textit{Stage 2: Reconstruction}}
   \STATE $G \leftarrow \text{BuildGraph}(E_{primitives})$
   \STATE $G' \leftarrow \text{MergeSplitBranches}(G)$ \COMMENT{Resolves V/L dummy nodes}
   \STATE $G'' \leftarrow \text{ResolveReferences}(G')$ \COMMENT{Resolves F/H/K dependencies}
   \STATE $N_{out} \leftarrow \text{AssignNamesAndValues}(G'')$
   \STATE \textbf{return} $N_{out}$
\end{algorithmic}
\end{algorithm}

\clearpage
\section{Dataset}
\subsection{Can Vision Language Models accurately Parse Netlists from Circuit Schematic Images?}
\label{app:vlm-ckt2nl}
To assess SOTA Vision Language Models' ability to parse valid SPICE netlists from circuit schematic images, we sampled 25 schematic images randomly from our large corpus of extracted circuit schematic and tasked four prominent VLMs with generating the corresponding netlists and descriptions.

Table \ref{tab:vlm-model-comparison} summarizes the performance of these baselines. Our empirical analysis reveals that Gemini 3 significantly outperforms other candidates in netlist extraction accuracy. Notably, Gemini 3 was the only model capable of consistently generating accurate netlists while simultaneously producing high-fidelity descriptions of the circuit. Consequently, we selected Gemini 3 as the backbone for our dataset generation pipeline.

\begin{table}[H]
\centering
\caption{Performance comparison of VLM baselines on a random subset of circuit schematic images.}
\label{tab:vlm-model-comparison}
\begin{small}
\begin{sc}
\begin{tabular}{lcc}
\toprule
Model & Sample Size & \% Correct \\
\midrule
Gemini 3   & 38 & 97.4\% \\
GPT-5.2    & 37 & 86.5\% \\
Gemini 2.5 & 38 & 76.3\% \\
GPT-4o     & 37 & 51.4\% \\
\bottomrule
\end{tabular}
\end{sc}
\end{small}
\end{table}

\subsection{VLM Prompt for Generating Netlist and Description Pairs from Schematic Images}

\begin{figure}[ht]
    \centering
    \begin{tcolorbox}[colback=gray!5, colframe=gray!50, title=\textbf{VLM Prompt for Schematic Extraction}]
    \small \ttfamily 
    
    You get two images: (1) a circuit crop and (2) the full source PDF page.
    Use the source page for labels/context; extract every circuit in the crop.
    Topology only; no simulation commands. Use standard SPICE components; X for subcircuits.
    First line of the netlist MUST BE title. Nodes numbered sequentially from 0, only numerical node names. If node names are given in the image, add the mapping in comments of the netlist.
    
    Output a JSON list of objects with:
    "netlist" (string), "name" (list of possible circuit names), "desc" (brief description).
    If no circuit is present, return null for all fields.
    Ensure valid JSON.
    
    Output MUST NOT include source page specific identifiers (figure numbers, book names, etc.). So, no figure names/numbers in the name.
    If a component value is missing, use a sensible placeholder (e.g., 10k resistors, 10u capacitors, 1N4148 diodes, 2N2222 BJTs, TL081 opamps).
    No need to include .model statements, they will be added later.
    Include all components in the netlist, even implicit ones (e.g., voltage sources not shown but written as Vdd, Vin etc.).
    
    VERY IMPORTANT: The netlist should only be for the circuit(s) present in the circuit crop image. The source page image is only for fetching description/labels, NOT for extracting additional circuits. ONLY look at the circuit crop image for the actual circuit topology.
    
    Make netlist according to standard NGSPICE syntax. Ngspice element guide:
    A:XSPICE B:Behavioral C:Cap D:Diode E:VCVS F:CCCS G:VCCS H:CCVS I:Current Source J:JFET K:MutualL L:Inductor M:MOSFET N:VerilogA O:LossyTL P:CPL Q:BJT R:R S:VSW T:TL U:RC/XSPICE V:V W:ISW X:Subckt Y:TXL Z:MESFET
    
    When wires overlap, carefully check whether they are electrically connected or merely crossing. An overlap alone does not imply a connection; typically, a dot indicates a junction, while no dot means the wires simply pass over each other. Although there can be exceptions, so pay close attention.
    \end{tcolorbox}
    \caption{The system prompt used for the VLM to extract netlists from schematic images.}
    \label{fig:vlm_prompt}
\end{figure}

\clearpage
\subsection{Dataset Preparation}
\label{dataset-prep}
\begin{figure}[H] 
    \centering
    
    \begin{subfigure}[b]{0.48\textwidth}
        \centering
        \includegraphics[width=\linewidth]{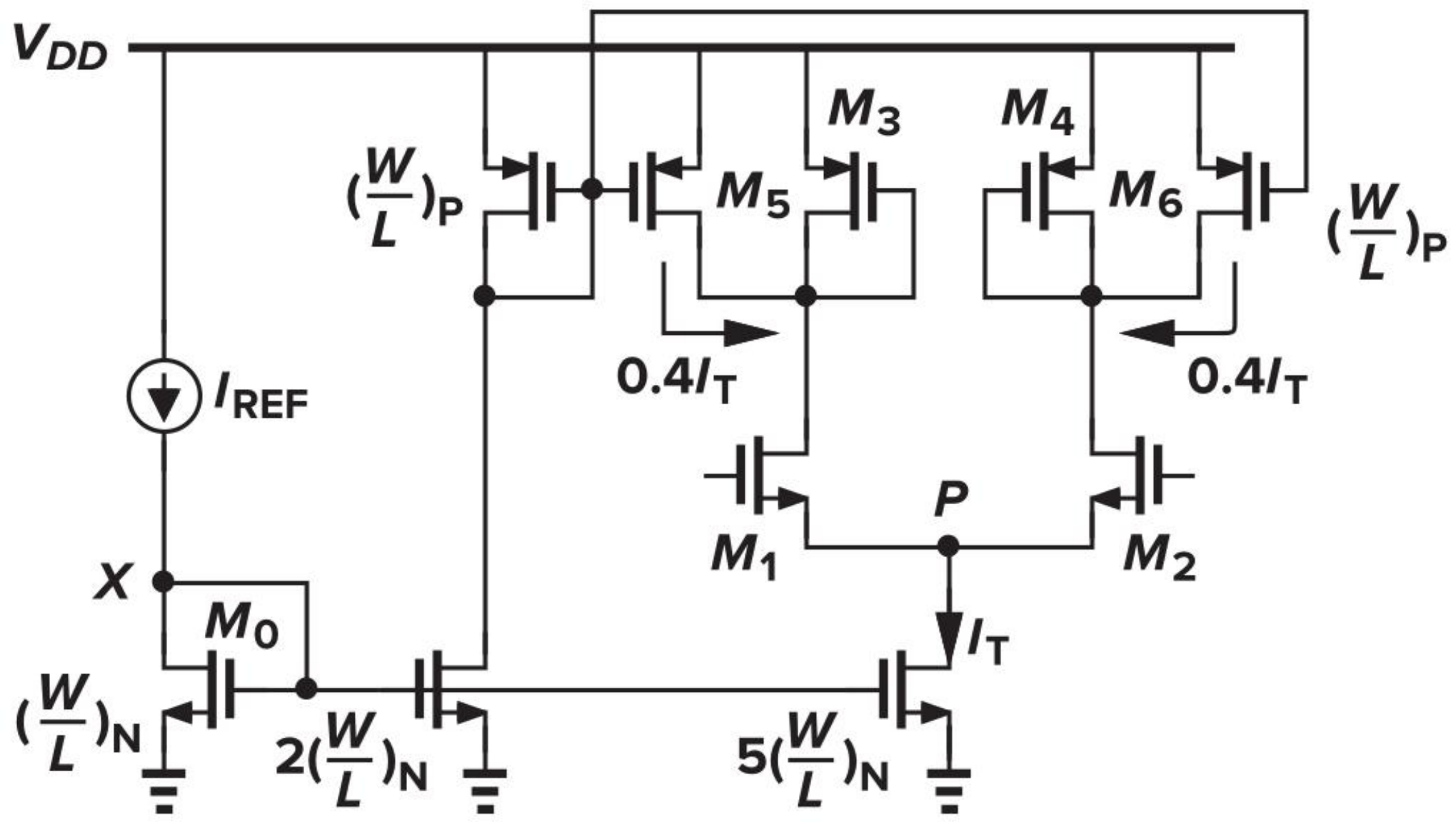}
        \caption{Circuit Schematic}
        \label{fig:schematic}
    \end{subfigure}
    \hfill
    \begin{subfigure}[b]{0.48\textwidth}
        \centering
        \includegraphics[width=0.7\linewidth]{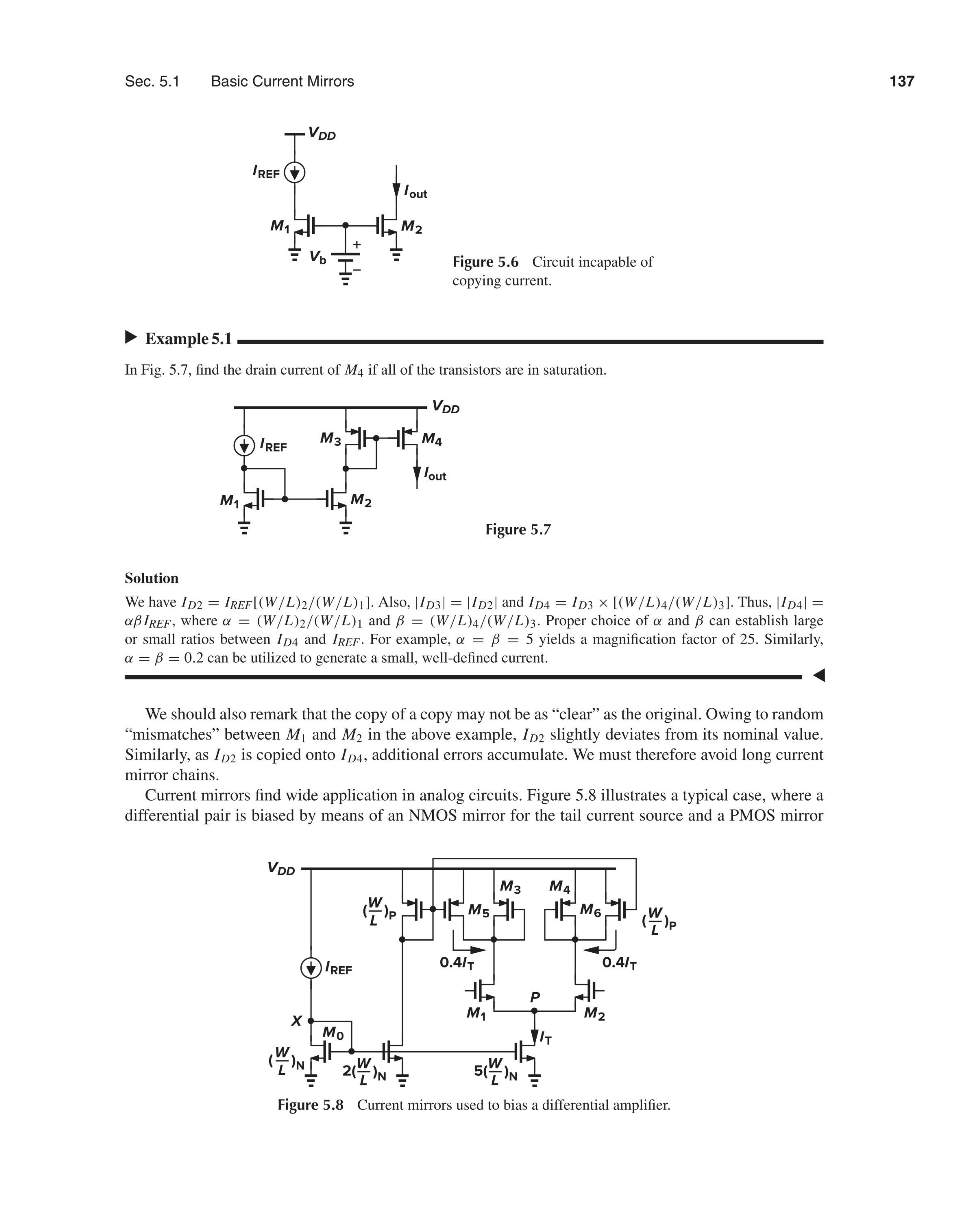}
        \caption{Source page for Context}
        \label{fig:context}
    \end{subfigure}
    
    \vspace{1em} 
    
    \begin{subfigure}[t]{0.48\textwidth}
        \tiny
        \textbf{Netlist for the circuit:}
\begin{verbatim}
Differential amplifier with current mirror biasing
.model NMOS NMOS
.model PMOS PMOS
Vdd 1 0 1.8
Iref 1 2 100u
* NMOS bias circuit
M0 2 2 0 0 NMOS W=10u L=1u
MB1 3 2 0 0 NMOS W=20u L=1u
MB2 4 2 0 0 NMOS W=50u L=1u
* PMOS bias circuit
MrefP 3 3 1 1 PMOS W=10u L=1u
* Differential amplifier stage
M1 5 7 4 0 NMOS W=20u L=1u
M2 6 8 4 0 NMOS W=20u L=1u
* Load and bias current sources
M3 5 5 1 1 PMOS W=10u L=1u
M5 5 3 1 1 PMOS W=10u L=1u
M4 6 6 1 1 PMOS W=10u L=1u
M6 6 3 1 1 PMOS W=10u L=1u
* Inputs
Vin1 7 0 0.9
Vin2 8 0 0.9
* Node mapping: VDD=1, X=2, bias_P=3, P=4, 
* Vout_L=5, Vout_R=6, Vin1=7, Vin2=8
\end{verbatim}
        \caption{Generated SPICE Netlist}
        \label{fig:netlist}
    \end{subfigure}
    \hfill
    \begin{subfigure}[t]{0.48\textwidth}
        \scriptsize
        \textbf{Description for the circuit:}
        
        "A differential amplifier stage where the tail current and the load currents are set by current mirrors from a single reference current source IREF. NMOS M0 and MB1-MB2 mirror the reference current to the tail node (P) and to a PMOS bias branch. PMOS MrefP then mirrors this current to M5 and M6, which act as current source loads in parallel with diode-connected PMOS loads M3 and M4."
        
        \vspace{1em}
        
        \textbf{Generated Names:}
        \begin{itemize}
            \item Current mirrors used to bias a differential amplifier
            \item Biased differential pair with active loads
            \item Differential amplifier with auxiliary PMOS current sources
        \end{itemize}
        \caption{Generated Description and Names}
        \label{fig:desc}
    \end{subfigure}

    \caption{Netlist and description pair correctly generated by VLM for circuit ID ckt\_33298469 in the dataset. (a) The extracted circuit schematic. (b) The original source page for VLM context. (c) The Ngspice compatible netlist generated for the schematic circuit. (d) The natural language description and names generated from the schematic.}
    \label{fig:dataset-prep-demo}
\end{figure}

Figure \ref{fig:dataset-prep-demo} shows how the dataset construction pipeline successfully generated a correct netlist and description from a given circuit schematic image.

\clearpage
\subsection{Dataset Statistics}
\label{app:dataset-stats}

\begin{figure*}[h]
    \centering
    \includegraphics[width=0.75\textwidth]{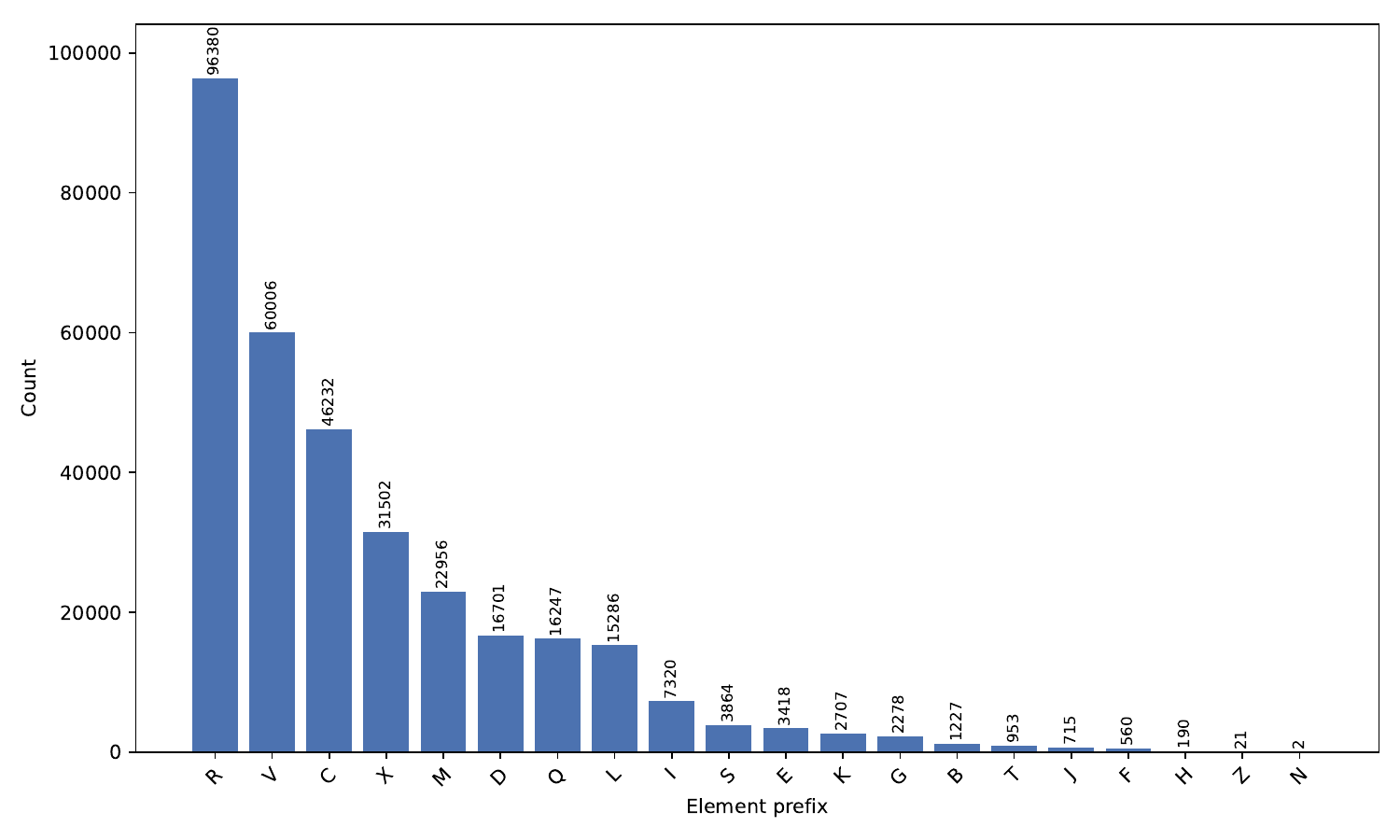}
    \caption{Distribution of SPICE elements in the dataset.}
    \label{fig:element_distribution}
\end{figure*}

\begin{figure*}[h]
    \centering
    \includegraphics[width=0.75\textwidth]{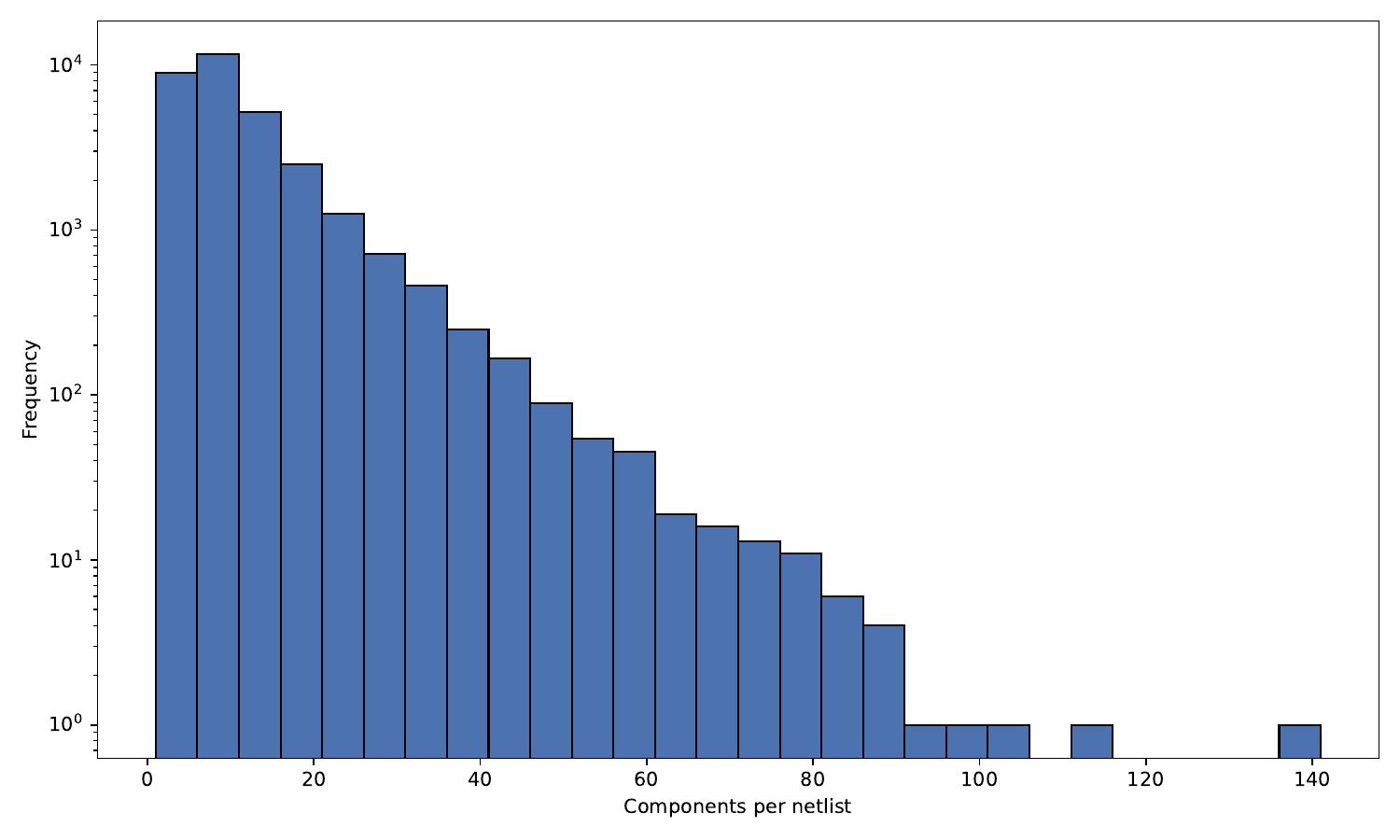}
    \caption{Histogram of netlist sizes in the dataset. Frequency axis is logarithmic.}
    \label{fig:netlist_size_histogram}
\end{figure*}

Figure \ref{fig:element_distribution} illustrates the distribution of circuit elements across the full dataset. Figure \ref{fig:netlist_size_histogram} presents the histogram of netlist sizes. Table \ref{tab:book_categories} summarizes the categories of all books processed to construct the dataset.

\begin{table}[ht]
\caption{Distribution of Source Books by Category Used in Dataset Construction.}
\label{tab:book_categories}
\begin{center}
\begin{small}
\begin{tabular}{lc}
\hline
Category & Count \\
\hline
Fundamental Circuit Analysis & 1 \\
Microelectronics Basics & 3 \\
General Analog IC Design & 4 \\
CMOS Design, Layout \& Modeling & 4 \\
Practical Electronics \& Troubleshooting & 3 \\
Operational Amplifiers (Theory \& App.) & 5 \\
Specialized Amplifiers (Opto/Feedback) & 2 \\
Audio Amplifiers \& Systems & 3 \\
Analog Filters & 2 \\
Data Converters (ADC/DAC) & 2 \\
Oscillators \& Frequency Control & 1 \\
RF General Design & 2 \\
RFIC Design (CMOS/Integrated) & 3 \\
Microwave Engineering \& Active Circuits & 3 \\
RF Passive, Lumped Elements \& Wireless & 2 \\
Power Electronics Principles & 4 \\
Switching Power Supply Design & 3 \\
Magnetics (Inductors/Transformers) & 1 \\
Sensor Technologies & 1 \\
Optoelectronic ICs & 1 \\
Interfacing \& Embedded Systems & 1 \\
Circuit Encyclopedias & 6 \\
Hobbyist Circuit Collections & 4 \\
Beginner/DIY Projects & 1 \\
\hline
\textbf{Total} & \textbf{62} \\
\hline
\end{tabular}
\end{small}
\end{center}
\end{table}

\clearpage

\newpage
\section{CircuitBench-100}
\label{sec:appendix-testset}
Table \ref{tab:test_set_stats} summarizes the circuit categories represented in the  CircuitBench-100 test set.

\begin{table} [H]
    \centering
    \caption{Distribution of Circuit Categories in the CircuitBench-100 Test Set}
    \label{tab:test_set_stats}
    \begin{tabular}{p{0.25\textwidth} p{0.55\textwidth} c}
        \toprule
        \textbf{Category} & \textbf{Constituent Topologies} & \textbf{Count} \\
        \midrule
        Amplifiers & Common-Emitter, Common-Source, Cascode, Differential, Op-Amps, Voltage Follower, Darlington, Output Stages, Microwave, JFET & 24 \\
        \addlinespace
        Basic Components \& Networks & Single Transistor Test, Diode Circuits, Resistive Bridges, RLC Basics, Parallel/Series Networks, Two-Port Networks & 18 \\
        \addlinespace
        Modeling \& Equivalent Circuits & Small/Large-Signal Models, Layout Models, Transformer Models, Resistor Models & 13 \\
        \addlinespace
        Biasing \& References & Current Mirrors, Widlar Source, Voltage Bias, Biasing Networks, Current Steering, Constant Current Sources & 12 \\
        \addlinespace
        Filters & Low-Pass, High-Pass, LC Ladder, Chebyshev, Butterworth, Pi-Network, Attenuators & 11 \\
        \addlinespace
        Power Electronics & Buck-Boost Converter, Switching Circuits, Rectifiers, Class E Amplifier & 6 \\
        \addlinespace
        Logic \& Digital & Logic Gates, Pass-Transistor Logic, Shift Registers, Transmission Gates, CMOS Inverter & 5 \\
        \addlinespace
        RF \& Systems & Transmission Lines (Kuroda), Y-Parameter Test, Feedback Systems, Control Blocks & 5 \\
        \addlinespace
        Oscillators & Colpitts, Cross-Coupled LC, RF Video Modulator & 3 \\
        \addlinespace
        Sensors \& Opto & Phototransistor, IR LED Driver, LED Indicators & 3 \\
        \midrule
        \textbf{Total} & & \textbf{100} \\
        \bottomrule
    \end{tabular}
\end{table}

\section{Model Architecture and Training Strategy}
\label{sec:model-architecture}

\textbf{CircuitFormer} is a hybrid encoder-decoder Transformer architecture \cite{NIPS2017_3f5ee243} designed to translate natural language specifications into valid circuit topologies. The architecture integrates a semantic-rich encoder with a syntax-aware decoder, bridged by cross-attention mechanisms. 

\subsection{The Hybrid Architecture}
\textbf{Encoder (Sentence-BERT):} Standard BERT embeddings often suffer from the anisotropy problem, where embeddings occupy a narrow cone in the vector space, limiting their utility for semantic similarity tasks. To mitigate this, we employ a pre-trained Sentence-BERT (sBERT)~\cite{reimers-gurevych-2019-sentence} as the encoder backbone. sBERT is optimized via siamese networks to derive semantically meaningful sentence embeddings, ensuring the model captures the nuances of the input prompts. The encoder utilizes a standard WordPiece tokenizer to process natural language.

\textbf{Decoder:} The decoder utilizes a GPT-2 architecture \cite{radford2019language} extended with cross-attention layers. Unlike the encoder, the decoder employs our custom {CKT Tokenizer} (Section \ref{ckt_tokenizer}) to efficiently represent circuit topologies. 

\textbf{Integration via Cross-Attention:} We integrate the encoder and decoder by injecting cross-attention layers into each block of the decoder. While the decoder's self-attention models intra-netlist dependencies, the cross-attention mechanism conditions the generation process on the sBERT encoder's semantic embeddings, effectively grounding the circuit generation in the natural language prompt.

\subsection{Three-Stage Optimization Strategy}
To effectively align the pre-trained modules and ensure robust performance, we adopt a three-stage training curriculum:

    \textbf{Stage 1: Domain-Adaptive Decoder Pre-training:} Before integration, we pre-train the decoder in isolation on our large-scale netlist corpus using a standard causal language modeling (CLM) objective. This step allows the model to internalize the SPICE netlist grammar and the topological structures encoded by the CKT tokenizer, independent of any natural language conditioning.
    
    \textbf{Stage 2: Alignment (Cross-Attention Warmup):} We then assemble the full architecture, initializing the cross-attention layers randomly while loading the pre-trained weights for the encoder and decoder. In this stage, we freeze the weights of both the sBERT encoder and the GPT-2 decoder, optimizing \textit{only} the newly initialized cross-attention layers. This phase aligns the decoder's latent space with the encoder's semantic representations without catastrophically forgetting the pre-learned syntactic features.
    
    \textbf{Stage 3: End-to-End Fine-Tuning:} Finally, we unfreeze the entire architecture and fine-tune all parameters in a supervised manner. This stage optimizes the global model to minimize the cross-entropy loss between the generated tokens and the ground-truth netlists, refining the interaction between semantic understanding and topological generation.

\section{The Judge LLM}
\label{app:judge-llm}
   The judge LLM evaluates functional equivalence by comparing the generated netlist against the prompt and the reference ground-truth topology. Using Gemini 3 as the judge, we performed a correlation study between automated LLM-based evaluations and human expert assessments of state-of-the-art models and CircuitFormer. As shown in Table \ref{tab:human-judge-corr}, the judge LLM shows strong agreement with human evaluations, supporting its use as a reliable and reproducible proxy for large-scale benchmarking.

 \begin{table}[h]
\centering
\caption{Comparison of success rates between Human Expert evaluation and the LLM Judge. The minimal difference validates the automated judge's reliability.}
\label{tab:human-judge-corr}
\begin{small} 
\begin{sc}
\begin{tabular}{l c c c}
\toprule
\textbf{Model} & \textbf{\shortstack{Manual\\Eval.}} & \textbf{\shortstack{Judge LLM\\Eval.}} & \textbf{\shortstack{Abs.\\Diff}} \\
\midrule
GPT-OSS & 69\% & 71\% & 2\% \\
Llama 3.3 & 31\% & 34\% & 3\% \\
Mistral Large 2 & 32\% & 29\% & 3\% \\
Gemma 3 & 23\% & 22\% & 1\% \\
DeepSeek V3 & 63\% & 65\% & 2\% \\
CircuitFormer & 83\% & 85\% & 2\% \\
\bottomrule
\end{tabular}
\end{sc}
\end{small}
\end{table}

\end{document}